\documentclass{article}
\usepackage{microtype}
\usepackage{graphicx}
\usepackage{subfigure}
\usepackage{booktabs} 
\usepackage{hyperref}

\usepackage[accepted]{icml2021}

\usepackage{multirow}
\usepackage{hyperref}

\usepackage{amsmath}

\icmltitlerunning{Class-Distribution-Aware Calibration for Long-Tailed Visual Recognition}

\begin{document}

\twocolumn[
\icmltitle{Class-Distribution-Aware Calibration for Long-Tailed Visual Recognition}


\begin{icmlauthorlist}
\icmlauthor{Mobarakol Islam}{icl}
\icmlauthor{Lalithkumar Seenivasan}{nus}
\icmlauthor{Hongliang Ren}{nus}
\icmlauthor{Ben Glocker}{icl}
\end{icmlauthorlist}

\icmlaffiliation{icl}{BioMedIA Group, Department of Computing, Imperial College London, UK}
\icmlaffiliation{nus}{Department of Biomedical Engineering, National University of Singapore, Singapore}

\icmlcorrespondingauthor{Mobarakol Islam}{m.islam20@imperial.ac.uk}

\icmlkeywords{Machine Learning, ICML}

\vskip 0.3in
]

\printAffiliationsAndNotice{} 

\begin{abstract}

Despite impressive accuracy, deep neural networks are often miscalibrated and tend to overly confident predictions. Recent techniques like temperature scaling (TS) and label smoothing (LS) show effectiveness in obtaining a well-calibrated model by smoothing logits and hard labels with scalar factors, respectively. However, the use of uniform TS or LS factor may not be optimal for calibrating models trained on a long-tailed dataset where the model produces overly confident probabilities for high-frequency classes. In this study, we propose class-distribution-aware TS (CDA-TS) and LS (CDA-LS) by incorporating class frequency information in model calibration in the context of long-tailed distribution. In CDA-TS, the scalar temperature value is replaced with the CDA temperature vector encoded with class frequency to compensate for the over-confidence. Similarly, CDA-LS uses a vector smoothing factor and flattens the hard labels according to their corresponding class distribution. We also integrate CDA optimal temperature vector with distillation loss, which reduces miscalibration in self-distillation (SD). We empirically show that class-distribution-aware TS and LS can accommodate the imbalanced data distribution yielding superior performance in both calibration error and predictive accuracy. We also observe that SD with an extremely imbalanced dataset is less effective in terms of calibration performance. Code is available in https://github.com/mobarakol/Class-Distribution-Aware-TS-LS.

\end{abstract}

\section{Introduction}
\label{introduction}
Modern deep neural networks have demonstrated the ability to achieve very high accuracy in recognition tasks. These deeper and wider networks can fit the training dataset with ease, achieving high accuracy~\cite{zhang2016understanding}. However, this raises a new concern, to what extend the network's predictions are likely to be correct? As these deep network tries to reduce the negative log-likelihood loss, they overfit to datasets, rendering its predictions to be over-confident and less trustworthy~\cite{mukhoti2020calibrating}. Here, the network is termed to be poorly calibrated. In a calibrated model, the accuracy matches the prediction confidence. A well-calibrated model is of utmost importance in a real-world application, at times assisting the user make a life-saving decision. For instance, letting the doctor decide whether to perform further diagnosis should the model's predictions be of low confidence. Network miscalibration is strongly related to the network's expanding capacity and the lack of regularization techniques~\cite{guo2017calibration}. Additionally, the long-tailed dataset has also been found to cause model miscalibration. As the dataset distribution is biased towards few very significant classes, in reducing the training loss, the model's predictions get over-confident towards these classes, leading to model miscalibration (see Appendix \ref{lt_dataset}, Figure \ref{fig:distribution}). This creates additional issues in calibrating probability prediction with conventional calibration methods.

The importance of model calibration over model accuracy has been acknowledged and worked upon recently. Most frequent methods used for model calibration are Platt scaling~\cite{platt1999probabilistic}, isotonic regression~\cite{niculescu2005predicting} and, bayesian binning and averaging~\cite{bella2009similarity}. Temperature scaling (TS), a variant of Platt scaling, has been proven to have a positive effect in calibrating the model~\cite{guo2017calibration}. In the temperature scaling technique, the network's confidence in its prediction is scaled down by diving the logits with a scalar value, called the temperature scale (T $>$ 0). The scaled logits help overcome the problem of overconfidence and thereby attaining better model calibration.

Most recently, label smoothing (LS) is introduced, a regularization technique to flatten the hard onehot label to improve accuracy, thereby implicitly reduces the overconfident predictions and demonstrate better confident calibration and feature representation~\cite{muller2019does}. A scalar smoothing factor controls the smoothing level where a higher value enforces a larger squeeze in the true label.

Despite the effective performance of TS and LS in probability calibration with the standard dataset, there are additional issues with a long-tailed dataset where the distribution of a few classes is significantly higher than the other. In reality, most sensitive applications do not possess a balanced dataset. Therefore, during training, the model gets biased towards reducing the loss of the few high-frequency classes, thereby causing over-confidence and miscalibration. In both techniques, TS and LS, a scalar smoothing factor applied uniformly over all the classes do not cater to probability prediction biases introduced by class distribution. In this work, we propose \textit{Class-Distribution-Aware} TS (CDA-TS) and LS (CDA-LS) that incorporates class frequency information to calibrate prediction probability in the long-tailed dataset.

The key contributions of this paper are as follows:(I) We introduce a class-distribution-aware temperature scaling (CDA-TS) and label smoothing (CDA-LS) techniques that improve model calibration despite training on the long-tailed datasets; (II) We design a knowledge-distillation (KD) loss with CDA-TS and demonstrate the superior performance of our KD loss in self-distillation; (III) We validate the effectiveness of our methods on three long-tailed datasets and four state-of-the-art probability calibration metrics, uncertainty calibration metric and reliability diagrams.

\section{Method}
\label{Method}

\subsection{Preliminaries}
\label{Background}
Most of the model calibration techniques try to reduce the over-confident prediction by manipulating either onehot label or predicted probability. For example, temperature scaling (TS) downgrades the logits by dividing them using a scalar temperature value and label smoothing (LS) squeezes the onehot label with a scalar smoothing factor. In TS, temperature scaled probability is $\hat{P}^{TS} = \sigma_{SM}(z/T)$, where temperature $T$ ($> 1$), logits $z$  and softmax layer $\sigma_{SM}$. To find the optimal temperature, a conjugate gradient solver or a naive line-search is used on a trained model without affecting model accuracy~\cite{guo2017calibration}. In LS, smoothened label or soft label is $P^{LS} = P^{OH}(1-\alpha) + \alpha/N$, where smoothing factor $\alpha$ ($0 < 1$), onehot label $P^{OH}$ and total number of classes $N$. The smoothing factor $\alpha$ controls the flatten degree in the smoothened label. A higher value provides a larger smoothened label. This is also called soft labels ($P^{LS}$) and used to calculate CE loss, $L_{CE} = \sum_{c=1}^{N} -{P^{LS}_c} log({\hat P_c)}$, where $\hat P_c)$ is the predicted probability. Though both TS and LS are proven to improve model calibration to an extend, model miscalibration is not entirely addressed since it can be affected by long-tailed distribution.

\subsection{Class-Distribution-Aware TS}
\label{Long-tail_TS}
TS addresses model miscalibration by dividing the logits with an optimal temperature value. By scaling the logits, TS restricts the model from being overly confident in its prediction. 
The TS scales all the logits using a scalar value uniformly. However, a model trained on a long-tailed dataset could be biased towards high-frequency classes, resulting in overly confident logits. There, we introduce a class-distribution-aware TS (CDA-TS) that incorporates class frequency information with the optimal temperature. We construct a vector temperature instead of a scalar by re-weighting the optimal temperature based on class frequency in the dataset. In this way, high-frequency classes divide with a larger value of the temperature and substantially downgrade the confidence score. 

If [$f_1, f_2,... f_N$] are the max-normalized class frequencies for N total classes and $T^{opt}$ is the optimal temperature then class-distribution-aware vector temperature $T^{cda}$ formulates is defined as:

\begin{equation} \label{eq:CDA_T}
    T^{cda} = T^{opt} + \gamma [f_1, f_2,..., f_N]
\end{equation}
where, $\gamma$ is the down-scale factor and the suitable value of $\gamma=0.1$ is found by the experiments. Therefore, the final equation of CDA temperature scaling is $\hat{P}^{CDA-TS} = \sigma_{SM}([z_1, z_2,..., z_N]/[T^{cda}_1,T^{cda}_2,...,T^{cda}_N])$. 

\begin{table*}[!h]
\centering
\caption{Performance of Class-aware TS on CIFAR-100-LT, Places-LT and ImageNet-LT dataset with ResNet18, ResNet152 and ResNet10, respectively.}
\scalebox{.88}{\label{table:cts}
\begin{tabular}{c|c|c|c|c|c|c|c|c}
\hline
Dataset & Ratio & Method  & ACC$\uparrow$ & ECE$\downarrow$ & SCE$\downarrow$ & TACE$\downarrow$ & BS$\downarrow$ & UCE$\downarrow$  \\ \hline
\multirow{6}{2.5cm}{CIFAR-100-LT} &\multirow{3}{1cm}{10} &Baseline &57.60 &0.1801 &0.0047  &0.0043 &0.6115 &0.2344 \\ \cline{3-9}
 & &TS (1.551) &57.60 &0.0425 &0.0034 &0.0035 &0.5656 &0.0502 \\ \cline{3-9} 
 & &CDA-TS &\textbf{57.93} &\textbf{0.0356} &\textbf{0.0032}  &\textbf{0.0033} &\textbf{0.5593} &\textbf{0.0305} \\ \cline{2-9} 
 
 &\multirow{3}{1cm}{100} &Baseline &38.57 &0.3280 &0.0082  &0.0071 &0.8976 &0.3892 \\ \cline{3-9}
 & &TS (2.0147) &38.57 &0.0372 &0.0047  &0.0056 &0.7580 &0.0194 \\ \cline{3-9} 
 & &CDA-TS &\textbf{38.90} &\textbf{0.0280} &\textbf{0.0046}  &\textbf{0.0055} &\textbf{0.7526} &\textbf{0.0099} \\ \hline
 
 \multirow{3}{2.1cm}{Places-LT} &\multirow{3}{1cm}{996} &Baseline &29.34 &0.2987 &0.0022  &0.0019 &0.9590 &0.4512 \\ \cline{3-9}
 & &TS (1.7349) &29.34 &0.0619 &\textbf{0.0014} &\textbf{0.0016} &0.8505 &0.1760 \\ \cline{3-9} 
 & &CDA-TS &\textbf{29.63} &\textbf{0.0581} &\textbf{0.0014} &\textbf{0.0016} &\textbf{0.8467} &\textbf{0.1689} \\ \hline
 
 \multirow{3}{2.1cm}{ImageNet-LT} &\multirow{3}{1cm}{256} &Baseline &34.83 &0.0853 &0.0007  &\textbf{0.0006} &0.7976 &0.2850 \\ \cline{3-9}
 & &TS (1.1200) &34.83 &0.0348 &\textbf{0.0006}  &\textbf{0.0006} &0.7897 &0.2301 \\ \cline{3-9} 
 & &CDA-TS &\textbf{35.68} &\textbf{0.0214} &\textbf{0.0006}  &\textbf{0.0006} &\textbf{0.7795} &\textbf{0.2116} \\ \hline
 
\end{tabular}}
\end{table*}

\begin{figure*}[!h]
    \centering
    \includegraphics[width=.9\textwidth]{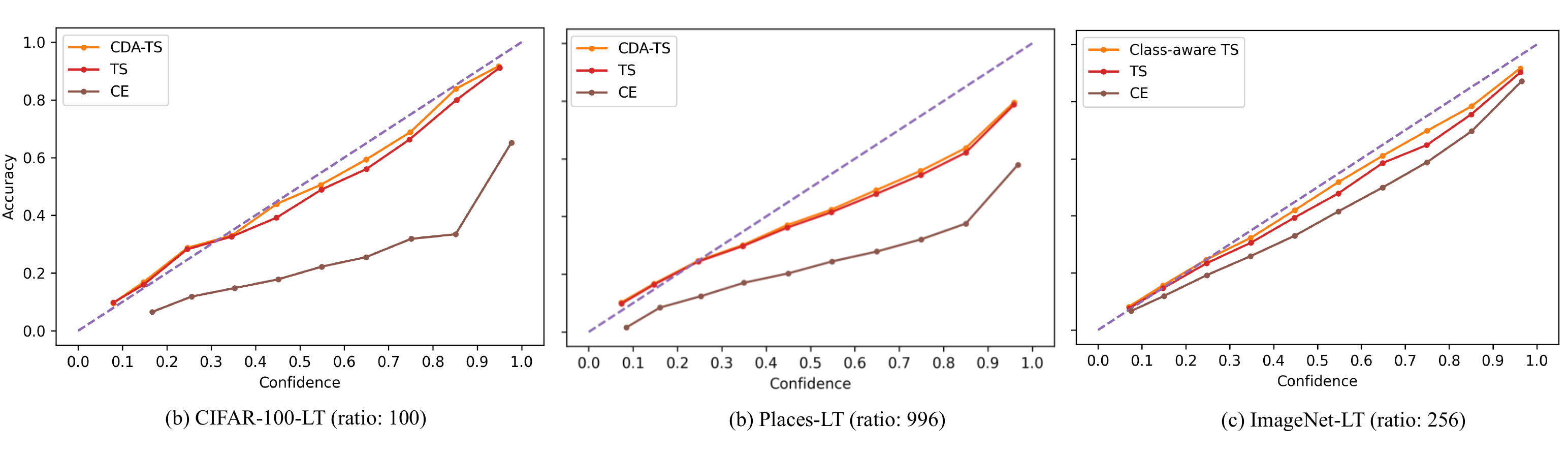}
    \caption{Reliability graph for CE, TS and CDA-TS with  (a) CIFAR-100-LT imbalanced ratio of 100, (b) Places-LT imbalanced ratio of 996 and (c) ImageNet-LT imbalanced ratio of 256.}
    \label{fig:reliability_plot_CDA_TS}
\end{figure*}

\subsection{Class-Distribution-Aware LS}
\label{Long-tail_LS}
Label smoothing (LS) limits the network to produce overly confident predictions by squeezing the true onehot label in CE loss calculation. In LS, onehot label flattens with a uniform distribution by using a smoothing factor. However, high-frequency classes in the long-tailed dataset overfit the model, limiting the LS performance with a uniform smoothing factor. To tackle this problem, we generate a class-distribution-aware LS (CDA-LS) soothing factor and replace the scalar smoothing factor in training with LS. CDA-LS flatten the high-frequency classes with a higher penalty than low-frequency classes. Similar to CDA-TS, the class-distribution-aware smoothing factor can be formulated as:

\begin{equation} \label{eq:CDA_T}
    \alpha^{cda} = \alpha + \gamma [f_1, f_2,..., f_N]
\end{equation}

where, the down-scale factor $\gamma=0.01$ is found by the experiments. Therefore, soft labels can be generated as, $P^{CDA-LS} = P^{OH}(1-[\alpha^{cda}_1, \alpha^{cda}_2,..., \alpha^{cda}_N]) + [\alpha^{cda}_1, \alpha^{cda}_2,..., \alpha^{cda}_N]/N$.

\subsection{Self-Distillation with CDA-TS}
\label{SD_Long-tail_LS}
A popular way to use knowledge distillation (KD) (as well as self-distillation(SD)) is to apply  Kullback-Leibler divergence loss on temperature (T) scaled logits of teacher and student model~\cite{hinton2015distilling}. A predefined value of T (e.g., $T=4$) is used to produces a softer probability distribution over the classes while calculating KD loss. Nevertheless, a long-tailed dataset produces overly confident probability to certain classes, and a fixed value of T is not tackling the long-tail effect during KD. Similar to CDA-TS, the information of class distribution can be integrated with the T in KD loss and formulated as-  

\begin{align}
\label{eq:dark_knowledge_distillation}
L_{SD}=
\sum_{c=1}^{N} \mathrm{KL}\Big(\sigma_{SM}\big(\frac{z^t(c)}{T^{cda}(c)}\big), \sigma_{SM}\big(\frac{z^s(c)}{T^{cda}(c)}\big)\Big) 
\end{align}

where $T^{cda}(c)$ is the CDA temperature for $c^{th}$ class and $z^t$, $z^s$ are the logits of teacher and student network, respectively. Student loss ($L_{CE}$) and distillation loss ($L_{SD}$) are fused to train the student model.

\section{Experiments}
\label{Results}

\subsection{Dataset}
\label{dataset}




\begin{figure}[!h]
    \centering
    \includegraphics[width=0.37\textwidth]{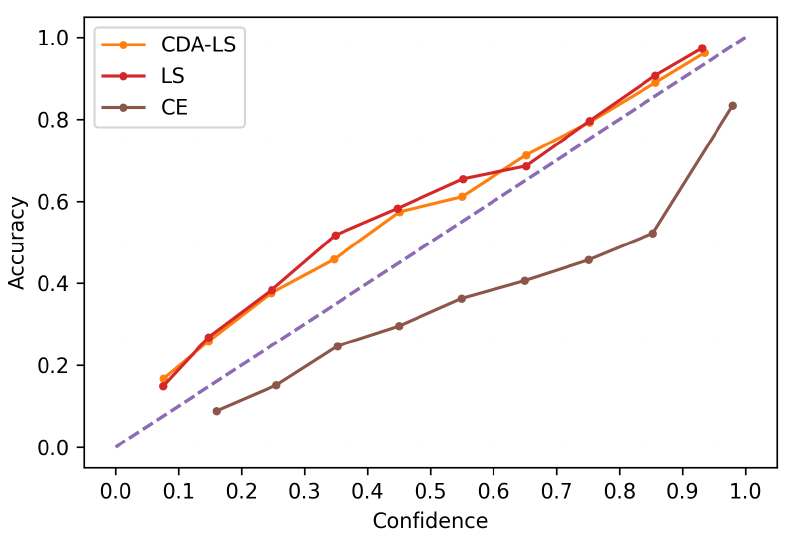}
    \caption{Reliability graph for CE, LS and class-aware LS with CIFAR-100-LT imbalanced ratio of 10.}
    \label{fig:reliability_plot_CDA_LS}
\end{figure}

\begin{figure}[!b]
    \centering
    \includegraphics[width=0.37\textwidth]{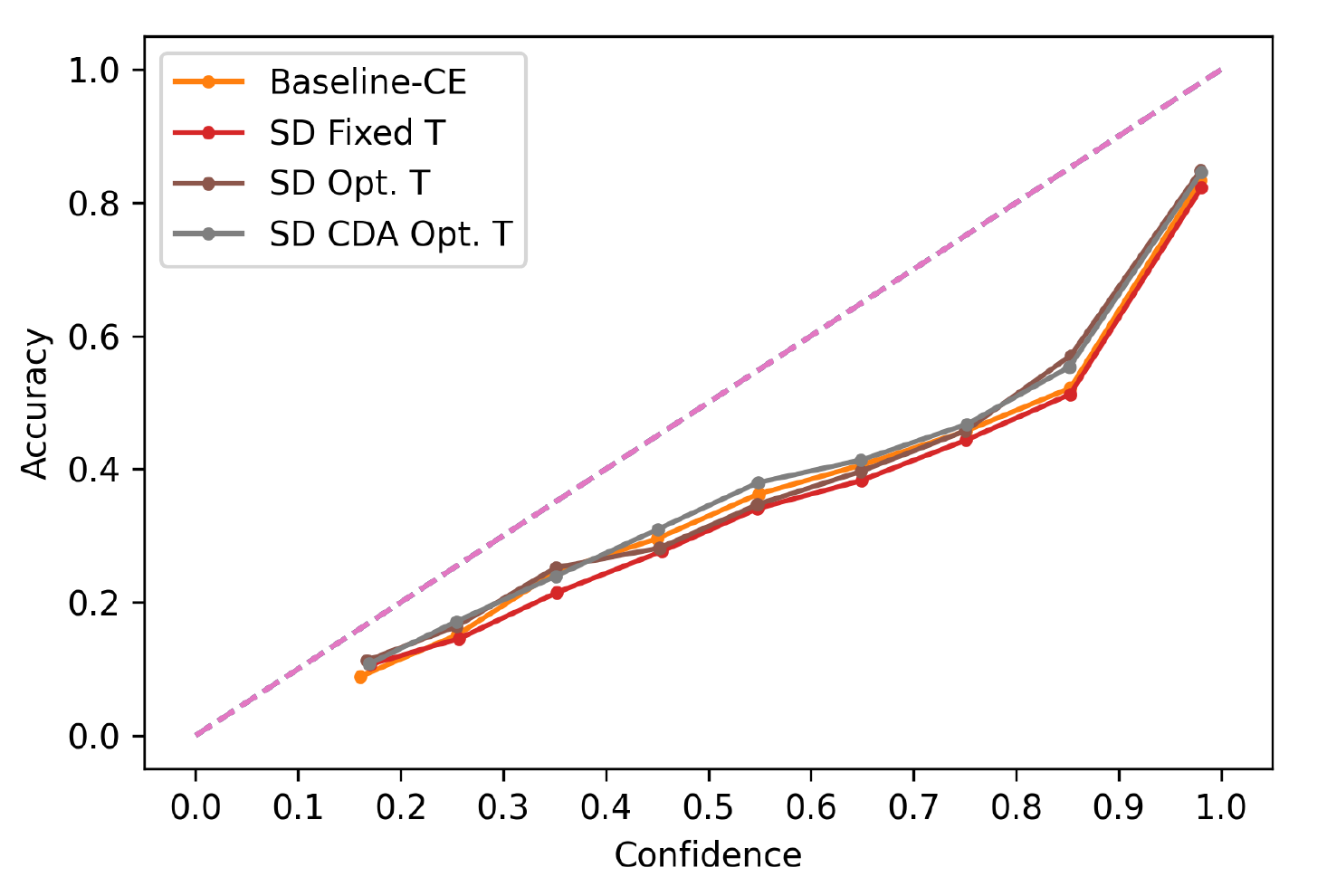}
    \caption{Reliability diagram for baseline-CE, SD with Fixed T and class-distribution-aware T with CIFAR-100-LT imbalanced ratio of 10.}
    \label{fig:reliability_sd_ts}
\end{figure}

The experiments are conducted on three long-tailed datasets: CIFAR-100-LT~\cite{krizhevsky2009learning}, Places-LT~\cite{zhou2017places} and ImageNet-LT~\cite{liu2019large}.  \textbf{CIFAR-100-LT}: The 100-class CIFAR-100 long-tailed dataset is generated by following~\cite{cao2019learning} with a data distribution imbalance factor of 10 and 100.  \textbf{Places-LT}: The 365-class Places long-tailed dataset with aa imbalance factor of 996 from~\cite{ ren2020balanced} is adopted. The dataset consists of $62.5$k training images and $36.5$k test images.  \textbf{ImageNet-LT}: similar to Places-LT dataset, the 1000-class ImageNet dataset with an imbalance factor of 256 is also adopted from~\cite{ ren2020balanced}. The dataset comprises $115$k training images and $20$k test images.

\subsection{Implementation}
The two CIFAR-100-LT with the 10 and 100 imbalance factor are trained on ResNet18~\cite{zhang2017mixup} backbone. The model is trained using an SGD optimizer with a batch size of 1024, a momentum of $0.9$, the number of train epochs of 200 and a learning rate of $0.1$. For Places-LT and ImageNet-LT dataset, pre-trained ResNet152 and ResNet18 models from~\cite{ ren2020balanced} are used, respectively. Cross-entropy (CE) loss is employed in training all the backbone networks.

\section{Evaluation}
The model’s performance and its miscalibration are quantified using accuracy (ACC) and various state-of-the-art calibration error metrics such as Expected calibration error (ECE)~\cite{guo2017calibration, naeini2015obtaining}, Static calibration error (SCE), thresholded adaptive calibration error (TACE), brier score (BS)~\cite{nixon2019measuring, ashukha2020pitfalls} and uncertainty calibration error (UCE) ~\cite{laves2019well} (see Appendix \ref{cal_metrics}).

As shown in Table \ref{table:cts}, the proposed CDA-TS is observed to outperform both the (a) baseline and (b) TS calibrated model and improve model calibration in all three long-tailed datasets. Reliability is also showing better calibration with CDA-TS in Figure \ref{fig:reliability_plot_CDA_TS}, bringing the confidence curve close to an identify function. The proposed CDA-LS is also observed to improve model calibration compared to baselines in Table \ref{table:cda_ls} and Figure \ref{fig:reliability_plot_CDA_LS}. On the other hand, SD based on CDA optimal T outperforms (a) base model and (b) SD based on fixed T~\cite{hinton2015distilling} in improving accuracy and model calibration (as shown in Table \ref{table:sd_ts}) and Figure \ref{fig:reliability_sd_ts}.





\begin{table}[!h]
\centering
\caption{Performance of class-aware LS on CIFAR-100-LT dataset. LS smoothing factor $\alpha=0.1$ is assigned for all the experiments.}
\scalebox{.90}{\label{table:cda_ls}
\begin{tabular}{c|c|c|c|c|c|c}
\hline
  & \multicolumn{3}{c|}{Imb. Ratio: 10} & \multicolumn{3}{|c}{Imb. Ratio: 100}  \\ \hline
 & ACC & ECE & UCE & ACC & ECE & UCE  \\ \hline
Baseline &57.60 &0.180 &0.234  &\textbf{38.57} &0.328 &0.389 \\ \hline
LS &57.98 &0.092 &0.153  &37.90 &0.045 &\textbf{0.040} \\ \hline
CDA-LS &\textbf{58.35} &\textbf{0.078} &\textbf{0.137} &37.94 &\textbf{0.038} &0.047 \\ \hline
\end{tabular}}
\end{table}

\begin{table}[!th]
\centering
\caption{CIFAR-100-LT performance of Class-distribution-aware TS on self-distillation with fixed temperature is set to 4 and optimal (opt.) temperature obtains 1.551 for the imbalanced ratio of 10. }
\scalebox{.90}{\label{table:sd_ts}
\begin{tabular}{c|c|c|c|c}
\hline
 Method  & ACC$\uparrow$ & ECE$\downarrow$ & TACE$\downarrow$ & UCE$\downarrow$  \\ \hline
Baseline &57.60 &0.1801  &0.0043 &0.2344 \\ \hline
SD (4) &57.64 &0.1974 &0.0044 &0.2498 \\ \hline
SD Opt. T (1.551) &57.95 &0.1714 &0.0042 &0.2264 \\ \hline
SD CDA Opt. T &\textbf{58.67} &\textbf{0.1694}  &\textbf{0.0042} &\textbf{0.2230} \\ \hline
 
\end{tabular}}
\end{table}





\section{Discussion and Conclusion}
\label{Discussion}
We have shown that the calibration of a model trained on a long-tailed dataset can be improved by employing class-distribution-aware temperature scaling (CDA-TS) and label-smoothing (CDA-TS). CDA-TS has shown improvement across accuracy and all calibration error matrix of models trained on CIFAR-100-LT, Places-LT and ImageNet-LT. While the proposed CDA-LS technique outperformed the traditional LS techniques in most of the metrics, the scalar LS techniques seem to perform better on other uncertainty calibration error metrics for high imbalance factor (as in Table \ref{table:cda_ls}). Furthermore, we show that the CDA-TS does not hinder knowledge distillation. It outperforms the baseline model and model trained using TS in accuracy (as in Table \ref{table:sd_ts}). It is observed that, while the use of TS generally increased calibration error when trained on highly imbalanced data, the proposed CDA-TS introduces lesser error compared to the TS technique.

Overall, we have shown that, by incorporating knowledge on class distribution, the CDA-TS and CDA-LS improve model calibration. Additionally, LS is known to hinders knowledge distillation~\cite{muller2019does, shen2021label}. Therefore, we further extend our work and show that CDA-TS improves accuracy and introduce lesser calibration error during knowledge distillation. In the future, we aim to study the effects of CDA calibration techniques in domain adaptation and introduce CDA loss function to improve model accuracy and calibration simultaneously. 

\section*{Acknowledgements}
This research has received funding from the European Research Council (ERC) under the European Union's Horizon 2020 research and innovation programme (grant agreement No 757173, project MIRA).

\nocite{langley00}

\bibliography{myreferences}
\bibliographystyle{icml2021}
\clearpage
\appendix
\section{Related Work}
\subsection{Network Calibration}
Several techniques are being introduced for calibrating networks trained on datasets with the less biased class distribution. The common methods are temperature scaling, isotonic regression, deep ensemble, label smoothing, and mixup. \citet{guo2017calibration} show that modern neural network is poorly calibrated, and temperature scaling (TS) is effective in calibration prediction. Variants of TS are also being explored. TS is extended to drop-out variational inference to produce well-calibrated model uncertainty~\cite{laves2019well}. Local temperature scaling is developed by focusing on calibration in multi-label semantic segmentation~\cite{ding2020local}. Other variations of TS include attended temperature scaling~\cite{mozafari2018attended}, bin-wise temperature scaling~\cite{ji2019bin} and TS with focal loss~\cite{mukhoti2020calibrating}. Dirichlet calibration, a multiclass calibration method, is introduced using Dirichlet distribution~\cite{kull2019beyond}. Recently, label smoothing presents as one of the efficient regularization techniques to improve the confidence calibration and feature representation~\cite{muller2019does, islam2020learning}.
\begin{figure}[!b]
    \centering
    \includegraphics[width=0.36\textwidth]{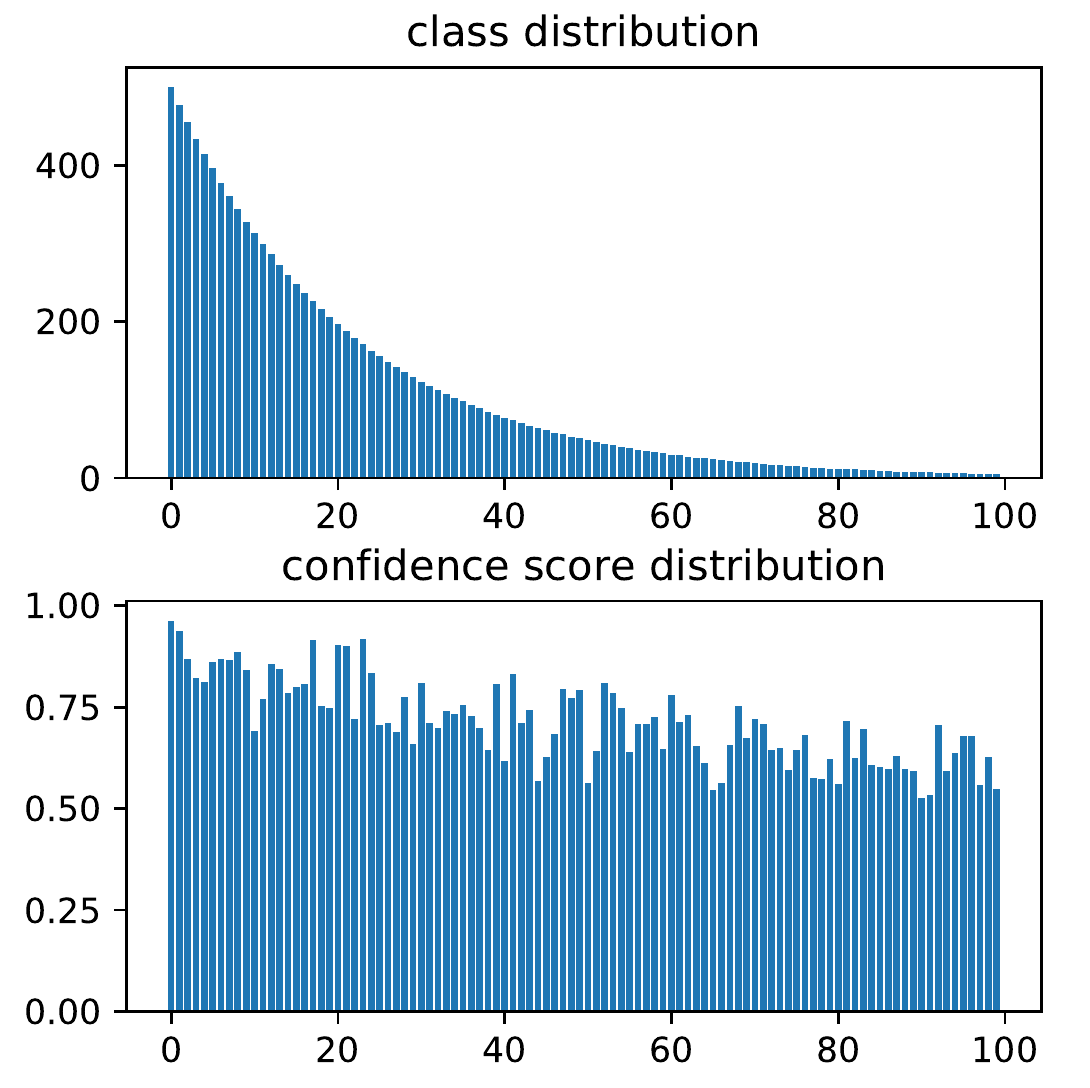}
    \caption{Relation of class-frequency-distribution and corresponding confidence score on CIFAR-100-LT.}
    \label{fig:distribution}
\end{figure}

\subsection{Long-Tailed Dataset}
\label{lt_dataset}
Long-tail is the common issue for pragmatic deep learning applications, consisting of extremely imbalanced class distribution in the training dataset (see Figure \ref{fig:distribution}). There are three general approaches to tackle long-tailed dataset, (a) re-sampling~\cite{byrd2019effect, buda2018systematic}, (b) Re-weighting~\cite{huang2019deep}, and (c) incorporating class frequency information~\cite{cao2019learning, ren2020balanced}. We are particularly interested in class frequency-based techniques in this work. Label distribution aware margin Loss is introduced to learn the imbalanced dataset in visual classification tasks~\cite{cao2019learning}. Balanced meta-softmax is designed with class distribution and meta-learning~\cite{ren2020balanced} where balanced group softmax trains the model group-wise to balance the classifier.

\subsection{Self-Distillation}
Self-distillation (SD) or knowledge distillation (KD) is a kind of transfer of knowledge from a teacher model to a student model by commonly minimizing an objective function~\cite{ba2013deep, crowley2018moonshine}. A popular way to do this is by calculating the Kullback-Leibler divergence loss between the temperature scaled probability of teacher and student models~\cite{hinton2015distilling}. Hardmax and softmax with temperature are used to perform self distillation for overparameterized neural network~\cite{dong2019distillation}.

\section{Calibration Metrics}
\label{cal_metrics}
The model’s performance and its miscalibration are quantified using various state-of-the-art calibration error metrics. Expected calibration error (ECE)~\cite{guo2017calibration, naeini2015obtaining} is used as the base calibration metric. ECE computes the discrepancy between accuracy and confidence probability. It can be formulate as, $ECE = \sum\limits_{b=1}^B \frac{n_{[b]}}{N}\mid Acc_{[b]} - Conf_{[b]}\mid$, where $B$, $N$, $n_[b]$, $Acc_{[b]}$ and $Conf_{[b]}$ are the total number of bins, total number of samples, and number of samples, accuracy and confidences in $b^{th}$ bin. However, it is indicated that ECE is sensitive to bin number and unable to present class-wise model calibration~\cite{nixon2019measuring}. Static calibration error (SCE), thresholded adaptive calibration error (TACE), brier score (BS) are introduced to deal with the limitation of ECE and determine accurate calibration error~\cite{nixon2019measuring, ashukha2020pitfalls}. TACE can be formulated as $TACE =\frac{1}{KR} \sum\limits_{k=1}^K \sum\limits_{r=1}^R \mid Acc_{[r,c]} - Conf_{[r,c]}\mid$, where $R$ is a number of calibration ranges with threshold t, $Acc_{[r,c]}$ and $Conf_{[r,c]}$ are the accuracy and confidence in r range and c class. 
We also calculate the uncertainty calibration error (UCE) by following~\cite{laves2019well}. The formula for UCE is similar to ECE except determining uncertainty calibration instead of probability calibration. Therefore, $UCE = \sum\limits_{b=1}^B \frac{n_{[b]}}{N}\mid Err_{[b]} - Uncert_{[b]}\mid$ where, total bin is B, number of samples in $b^{th}$ bin is $n_{[b]}$ and error and uncertainty in the corresponding bin are $Err_{[b]}$ and $Uncert_{[b]}$ respectively. For simplicity we assigned bin size of 10 for all the calibration metrics and threshold of $10^{-3}$ for TACE. The reliability diagram is used to visualize the model’s miscalibration for under-confidence or over-confidence.

\end{document}